# Target specification bias, counterfactual prediction, and algorithmic fairness in healthcare

Eran Tal

McGill University

eran.tal@mcgill.ca

**Abstract:** Bias in applications of machine learning (ML) to healthcare is usually attributed to unrepresentative or incomplete data, or to underlying health disparities. This article identifies a more pervasive source of bias that affects the clinical utility of ML-enabled prediction tools: target specification bias. Target specification bias arises when the operationalization of the target variable does not match its definition by decision makers. The mismatch is often subtle, and stems from the fact that decision makers are typically interested in predicting the outcomes of counterfactual, rather than actual, healthcare scenarios. Target specification bias persists independently of data limitations and health disparities. When left uncorrected, it gives rise to an overestimation of predictive accuracy, to inefficient utilization of medical resources, and to suboptimal decisions that can harm patients. Recent work in metrology – the science of measurement – suggests ways of counteracting target specification bias and avoiding its harmful consequences.

**ACM Reference Format:**

Eran Tal. 2023. Target specification bias, counterfactual prediction, and algorithmic fairness in healthcare. In *Proceedings of the 2023 AAAI/ACM Conference on AI, Ethics, and Society (AIES'23), August 8-10, 2023, Montreal, Canada.* ACM, New York, NY, USA, 10 pages. https://doi.org/10.1145/3600211.3604678

## 1 Introduction

Supervised machine learning (ML) is an increasingly common methodology for training models that support medical tasks such as diagnosis, treatment planning, and resource allocation. A growing body of research addresses the biases associated with such models and the impact of their use on the fairness and safety of medical decision making [1,5,9,13,15,16,32,34,36,37,39]. Currently, there is no consensus on how such biases should be reported to decision makers, e.g., to medical staff who prioritize hospital beds or refer patients for diagnostic tests. Particularly, it is unclear whether and how the presence of biases should affect the estimated *accuracy* of model outputs that is reported to medical staff. The literature on bias and fairness in ML tends to treat bias and accuracy as orthogonal properties of a model, and to allow the possibility that a given model is highly accurate but deeply biased, and vice versa [29,35]. This is consistent with the technical, probabilistic definitions of bias and accuracy accepted by ML researchers. And yet, from the perspective of a typical healthcare professional, these technical definitions are obscure and counterintuitive. Healthcare professionals take their understanding of bias and accuracy from medical measurement: when a blood test or echocardiography is biased, it suffers from a systematic measurement error, and is therefore inaccurate.

The tension in the meanings of terms like 'accuracy' and 'bias' between measurement and ML is not merely a terminological issue. Instead, it is emblematic of a mutual misunderstanding of how medical professionals think about the targets of prediction versus the way algorithm designers operationalize target variables. If not addressed, this misunderstanding can give rise to misinterpretation of model outputs and to suboptimal decisions that are harmful to patients. In what follows, I propose a way of conceptualizing and communicating the accuracy of ML-based decision support tools that is in line with medical expectations and reduces health risks due to interpretive gaps between algorithm designers and users.

The accuracy of ML-based medical decision support tools depends on two broad factors: the predictive accuracy of the model, and the accuracy of the benchmarks against which model accuracy is evaluated. Sources of inaccuracy that fall under the first factor include under- and over-fitting, unrepresentative or small datasets, and imbalanced datasets, among many others. This article focuses on the second factor, namely, the accuracy of the benchmarks used to evaluate the accuracy of ML models. In supervised ML, these benchmarks are usually taken to be the labels in the validation and test datasets. Accordingly, significant efforts to improve accuracy in medical ML decision support tools have concentrated on improving the quality of labels [2,40,44].

While these efforts are important and laudable, this article highlights another source of benchmark inaccuracy in medical decision support tools that cannot be remedied simply by improving the quality of labels, and persists even in the hypothetical case where labels perfectly reflect the medical reality underlying the data. This additional source of benchmark inaccuracy is *target specification bias*. As its name suggests, this kind of bias arises due to differences between the way the target variable is specified from the perspective of medical decision makers, and the way the target variable is operationalized by the labels in the validation and test datasets. As I will show, a common source of target specification bias is that medical decision makers are typically interested in predicting variables that are specified under counterfactual conditions, while labels can only operationalize those same variables under actual conditions. As a result, labels may be biased with respect to the target variable even when the labels are reliably obtained and carefully curated.

I borrow the theoretical framework for the concept of target specification bias from *metrology*, the science of measurement. A central goal of metrology is to supply universal and replicable benchmarks for evaluating measurement accuracy, such as the standard metre, kilogram and second. I will contrast the modern concept of metrological accuracy with the 'label-matching' concept of accuracy currently prevalent in the literature on supervised ML, and find the latter lacking for the purposes of reporting to decision makers. I will then propose a broader concept of benchmark accuracy for medical ML decision support tools that is inspired by metrology. This broader concept of benchmark accuracy takes into account not only label quality, but also target specification bias.

Target specification bias is closely tied to fairness. The counterfactual scenarios under which medical decision makers typically specify their target variables are also the ones they use to define what counts as a fair decision. This is consistent with counterfactual conceptions of fairness in ML [6,28]. I will conclude by arguing that substantive considerations concerning fairness and the dynamics of healthcare decision making are intrinsic to specifying benchmarks for model accuracy. The accuracy of ML decision support tools in medicine should be reported relative to such benchmarks, rather than merely based on their label-matching rates. Doing so would increase the fairness and safety of such tools.

## 2    The label-matching conception of accuracy

Several measures of accuracy are used in the machine learning literature. The most straightforward one is the probability of a match between a model's predictions and the values of the target variable, $p(Y = \hat{Y})$. Other, more sophisticated measures of accuracy, such as the area under the ROC curve, are functions of the probabilities of a match or mismatch between predictions and target variable values. Determining the accuracy of a machine learning model thus requires an estimation of the values of the target variable. The common practice in supervised machine learning is to estimate target variable values from labels in the test dataset. These labels are produced by a source that is assumed to be reliable. In medicine, labels for diagnoses are commonly produced by 'gold standards' of evidence. These may include the verdict of a pathologist based on an analysis of a biological sample. For example, labels in a screening tool for skin cancer are produced by pathologists who examine the results of a biopsy of a skin lesion [1,12]. The predictions of the screening algorithm are deemed accurate to the extent that they replicate biopsy results.

The machine learning research community is well aware that labels may be inaccurate, and that training datasets may misrepresent the target population. Much recent attention has been given to overcoming 'label bias' and data acquisition error (or 'measurement error') and to diversifying commonly used training datasets [2,40,44]. These efforts increase benchmark accuracy, and with it the reliability of evaluations of predictive accuracy. However, the question remains as to whether a model that replicates the labels in a reliable and representative dataset should for that reason be deemed accurate. Current practice suggests that much of the machine learning community assumes that the answer is 'yes'. A look at recent reports of machine learning applications in medicine shows that researchers consistently select measures of accuracy that strictly track the replication of labels [10,14,20,25,27]. By 'strictly track' I mean that the model's accuracy is evaluated as a monotonically increasing function of the probability of a match between predictions and labels in the test dataset, and that the model is considered 100% accurate if and only if its outputs match all the labels in the test dataset.

Should evaluations of algorithmic accuracy strictly track the match between predictions and labels? In what follows, I will call the view that a machine learning model is predictively accurate to the extent that its predictions match the labels in a reliably obtained and representative dataset the 'label-matching conception of accuracy'. The label-matching conception takes labels in a reliable and representative dataset to be unbiased operationalizations of the target variable. For example, if the target variable is the risk of cancer associated with a skin lesion, and the labels are biopsy results, the target variable is operationalized by the probability that a lesion with similar features would result in a positive biopsy result. As long as the dataset is a reliable representation of medical diagnostic practice, the label-matching conception assumes that the probability of a positive diagnosis tracks the risk of skin cancer.

The main advantage of the label-matching conception of accuracy is its simplicity. As long as the dataset is of sufficiently high quality, evaluating accuracy is simply a matter of counting matches (or calculating distances) between predictions and labels, and applying one of several mathematical transformations to the results. No additional information or expertise concerning the target problem domain, e.g., dermatopathology, is required to evaluate accuracy. On the other hand, the disadvantage of this conception is that it runs the risk of operationalizing the target variable in a manner that misaligns with the way users and stakeholders define it. For example, there are reasons to doubt that the occurrence of positive biopsy results is an adequate operationalization of the occurrence of skin cancer. Due to racial disparities in the diagnostic process, Black patients in the US are typically diagnosed at a later disease stage. Hence, samples from early-stage Black patients may be underrepresented in the data [19]. In other words, while biopsy may be highly reliable in detecting skin cancer, and while the dataset may be representative of diagnostic practice, diagnostic practice is not directly reflective of the target variable that decision-makers and stakeholders are interested in predicting. Decision makers are interested in the occurrence of skin cancer, and are thus interested in predicting the diagnosis that a patient with a given set of features *would have received* had diagnostic practice been equally reliable for Black and white patients. The correct way to operationalize the target variable from the perspective of decision makers is as a predictor of a counterfactual scenario, rather than the actual scenario the model is trained to predict.

There are several reasons why actual medical practice may give rise to data that, despite being reliable, cannot be used to directly track the variable of interest. Not all such reasons involve a disparity that needs to be corrected, such as the late diagnosis of skin cancer in Black patients. In many cases, the mismatch between labels and target variable arises precisely because medical practice proceeds correctly and responsibly. Brian Christian discusses an early example of such mismatch [8]. In 1995, computer scientist Rich Caruana, then a graduate student at Carnegie Mellon University, was part of a team developing machine-learning algorithms to inform hospital admission decisions for pneumonia patients. The goal was to help physicians decide which pneumonia patients to hospitalize and which to follow up as outpatients. Caruana used patients' health outcomes, and specifically patient mortality, as labels. The neural net he trained achieved good accuracy (AUC=0.86) in predicting patient mortality [4]. However, a rule-based learning algorithm that was trained on the same data learned the rule that having asthma *lowers* the risk of mortality of a pneumonia patient. Conversations between the computer scientists and physicians revealed the likely cause: physicians tended to direct more resources to treating pneumonia patients with a known history of asthma. For example, such patients were often admitted directly into the intensive care unit, where they received aggressive care. This reduced the mortality rate of asthmatics with pneumonia relative to the overall pneumonia patient population.

Ironically, this meant that these patients were deemed low-risk by Caruana's model, which was trained on the same data and was accurate in predicting mortality.

It is worth examining precisely what went wrong with Caruana's model. The model's target variable – the thing it was designed to predict – was patient mortality. Risk of patient mortality was specified as the deciding factor for priority of hospital admittance as an inpatient. The labels used to train and test the model were records of patient mortality, and there is no reason to think that the labels were plagued by significant data acquisition error, i.e., that deaths were miscounted. Prima facie, then, the labels seemed to be good representations of the target variable. On a closer look, there was a misalignment between what the labels represented and the intended target variable. While the labels were records of actual mortality, the intended target variable was *ceteris paribus* mortality, that is, mortality 'all other things being equal'.

The variable physicians were interested in estimating when making decisions about inpatient admittance was not whether a given patient would in fact die of pneumonia. Whether or not a patient dies of pneumonia depends on other factors besides their health state and health-related risk factors. Specifically, it also depends on the quality and timeliness of the medical treatment they receive. The target variable of interest, rather, was the risk of death from pneumonia under a counterfactual scenario where all patients receive the same quality of care. Only under such counterfactual scenario can physicians control for the impact of care on a patient's health outcomes. The target variable is therefore *idealized*, in the sense that it represents a simplified and unrealistic scenario. The labels used to train and test Caruana's model reflected a real, complex scenario that was an imperfect approximation of the idealized, *ceteris paribus* case. These imperfections, if not detected in time, would have caused harm to asthmatics, who would have been de-prioritized for inpatient care had they contracted pneumonia.

## 3   Decision makers care about counterfactual prediction

The cases discussed above suggest that the label-matching conception of accuracy is misaligned with the interests of decision makers and other stakeholders. Even when the data are representative of the actual world, and the model is generalizable to other, real-world cases not included in the training data, model predictions may still be inadequate as operationalizations of the target variables decision makers care about. Figure 1 illustrates this mismatch for a simplified use case of ML in healthcare. The use case concerns decisions about treatment based on a prediction of a patient's health outcomes. In a simplified causal model of the underlying data generation process, two variables affect a patient's health outcomes (O): the characteristics (features) of the patient prior to medical intervention (X), and the characteristics of the healthcare interventions that the patient undergoes (I). This causal model is simplified inasmuch as other, e.g., environmental and social factors also contribute to health outcomes. Nonetheless, the simplified causal model is sufficient to demonstrate the mismatch between operationalization and definition that gives rise to target specification bias.

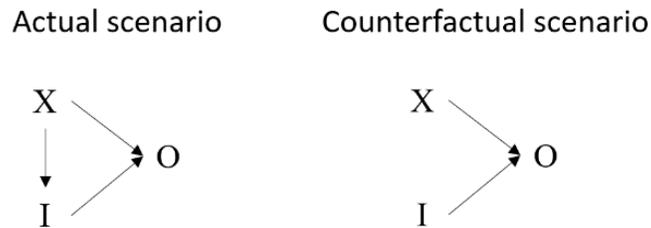

**Figure 1: A simplified causal model of data generation for health outcome prediction. X stands for patient features prior to medical interventions, I stands for the characteristics of medical interventions, and O stands for the patient's health outcomes. Actual data are generated by the scenario on the left, whereas decision makers are typically interested in predicting outcomes under the scenario on the right.**

Two scenarios of the simplified causal model may be distinguished. In an actual scenario, the characteristics of the healthcare interventions that a patient undergoes are affected by the features of the patient. For example, different patients that present symptoms of a similar kind are often treated differently based on the severity of their symptoms, their age, and their medical history. Moreover, the same intervention may be more or less effective depending on a patient's age, sex, and background medical conditions. Some patients refuse certain treatments, and this again may be correlated with the patient's age, sex, religion, and health condition. Finally, patients belonging to different socio-economic, racial, or ethnic groups often do not have the same range and quality of medical interventions available to them. As a result, the demographic and medical characteristics of a patient affect the type, duration, and quality of the medical interventions they undergo. Any accurate prediction of actual health outcomes is, by virtue of being accurate, necessarily sensitive to the interaction between patient features and the characteristics of the intervention.

To physicians who are tasked with making decisions about treatment, this interaction is a confounding factor. Physicians are typically interested in the difference a given intervention would make to the health of a patient, such as reducing their risk of death or increasing their quality of life. For this evaluation to be possible, physicians need information on how different patients are likely to respond to the *same* intervention. In other words, physicians are interested in the counterfactual scenario under which patient features and the characteristics of the intervention do not interact. For example, they are interested in the health outcomes different patients with pneumonia would have had 'all other things being equal', that is, if they had been given the same treatment. As Jessica Paulus and David Kent put it, "observed mortality is an imperfect proxy for mortality under ideal care, the true outcome of interest when constructing models for [medical] futility" [34:5].

Depending on the number of treatment options under consideration, physicians may be interested in multiple counterfactual predictions. For example, they may be interested in comparing the counterfactual health outcomes of a patient in a given population where all patients receive the same treatment T1 to their health outcomes in a population where all patients receive the same treatment T2, or perhaps no treatment at all. Only relative to such counterfactual scenarios can decision makers isolate the contribution of patient features to health outcomes, and select interventions accordingly.

A similar point holds for diagnostic use cases. Under a simplified causal model, patient characteristics and the characteristics of the diagnostic procedure both affect the diagnosis a patient receives. In an actual scenario, the features of a patient affect the characteristics of the diagnostic procedure they receive. Different patients presenting similar symptoms are often offered different diagnostic procedures depending on their age, sex, and whether or not they are known to belong to certain risk groups. The same diagnostic procedure may have varying sensitivity and specificity for different patients depending on age, sex, genetic profile, background health conditions, and a variety of other factors. The proportion of patients who refuse to undergo a certain diagnostic procedure may vary in correlation with demographic properties. And patients belonging to different socio-economic, racial, or ethnic groups often do not have the same range and quality of diagnostic procedures available to them.

Here again, the demographic and medical characteristics of a patient affect the type, quality, and timeliness of the diagnostic procedure they will undergo. A machine-learning model that accurately predicts actual diagnoses is necessarily sensitive to the interaction between patient features and the characteristics of diagnostic procedures. However, from the point of view of a physician who is required to decide whether or not to refer a patient to a diagnostic test, this interaction is a confounder. Such a physician is interested in evaluating the likelihood that the test would reveal important information, e.g., confirm or rule out the presence of a medical condition. As part of this evaluation, physicians are interested in determining a patient's risk of developing a given medical condition, regardless of whether or when that condition will, as a matter of fact, be diagnosed. In other words, physicians are interested in predicting the diagnosis the patient *would* receive in a counterfactual scenario where all patients with a given set of symptoms undergo a timely and accurate diagnosis procedure.

The comparison of actual and counterfactual scenarios demonstrates the limitations of the label-matching conception of accuracy. The ability of an ML model to reproduce labels in a generalizable way is an important step toward clinical utility, but is not sufficient. Labels in the dataset reflect health outcomes (or diagnoses) obtained in actual healthcare scenarios, whereas decision makers typically require information about counterfactual scenarios where some background causal factors are held fixed. To be clinically useful, an ML model must predict the health outcomes (or diagnoses) associated with patient features under such counterfactual, *ceteris paribus* scenarios. These counterfactual health outcomes (or diagnoses) differ from the labels in the dataset, not due to any measurement error, but because real data includes correlations that confound the relationship between patient features and health outcomes (or diagnoses) that decision makers are interested in learning about. As a result, when the accuracy of an ML model is evaluated based on the model's ability to reproduce labels, it is evaluated against a different variable than the one decision makers typically care about [34,36].

From the point of view of decision makers, then, the label-matching conception overestimates the accuracy of model predictions. Specifically, it does not account for discrepancies between the values of the target variable as it is operationalized by labels, and values of the target variable as it is defined by decision makers. Such discrepancies constitute target specification bias.

In response to this charge, one could argue that machine learning is not designed to predict counterfactual scenarios. Rather, a machine learning model is deemed accurate when there is a good fit between the associations learned by the model and the associations found in the real world. As long as the labels in the dataset are accurate and reliable representations of the target variable of interest, and the model generalizes well from the training dataset to new examples, the model should be deemed accurate for the patient population from which the data was collected.

This objection, though plausible at first, rests on a confusion between intended labels and target variables. Intended labels are the labels a dataset would have if the data were representative and reliably collected. Examples are cancer diagnoses, records of hospital admission, and records of death. However, even the labels in an ideally collected and curated dataset are not values of the target variable. Rather, they are *operationalizations* of the target variable, that is, empirically accessible stand-ins for the values of the target variable. The target of prediction in most applications of machine learning is a *latent variable*, that is, a variable whose values are not directly accessible through any empirical procedure, but require inference from available data [21]. To access their values, target variables must be operationalized. Whether or not labels are an adequate operationalization of the target variable depends on the definition of the target variable, and on the validity of the inference from labels to target variable values. The need for such inferences and their complexity have long been recognized in sciences that specialize in measurement, such as metrology and psychometrics. In the next section, I turn to an example from metrology, and examine how the accuracy of measuring instruments is evaluated in the face of gaps between the desired target variable and its operationalizations. I then build on this example in the following section, where I define target specification bias and offer ways of mitigating it.

## 4   A metrological conception of accuracy

Metrology, the science of measurement, is concerned with the practical and theoretical aspects of measuring. Metrologists are typically physicists and engineers who design and calibrate highly accurate measuring instruments, maintain and improve measurement standards, and regulate national and international systems of measurement, including the International System of Units (SI). While its orientation is mostly applied, metrology has also generated a considerable body of conceptual and methodological work. The *International Vocabulary of Metrology* (VIM), for example, discusses the meanings of general terms such as 'measurement accuracy', 'measurement error' and 'measurement uncertainty' [22]. Similarly, the *Evaluation of Measurement Data – Guide to the Expression of Uncertainty in Measurement* (GUM) provides a wealth of concepts and methods for evaluating measurement uncertainty [26].

Metrology provides valuable conceptual tools for judging the adequacy of an operationalization of a variable. In metrology, the quantity intended to be measured is called a 'measurand' [22:2.3]. The task of defining a measurand is distinguished from the task of realizing it. The distinction between realization and definition of measurands is central to modern metrology, and a key to its success in delivering reproducible measurement results. The *definition* of a measurand is a linguistic entity that specifies the conditions under which the quantity is intended to be measured. These conditions are often ideal and not obtainable in practice. For example, the standard unit of time, the SI second, is defined as the duration of exactly 9,192,631,770 periods of the electromagnetic radiation corresponding to the transition between two hyperfine levels of the unperturbed ground state of the cesium-133 atom [3:2.3.1]. The cesium atom in question is assumed to be unaffected by gravitational fields, magnetic fields, or thermal radiation, and to have no interactions with other atoms. These are counterfactual conditions that cannot be practically achieved in a laboratory.

The definition of the SI second assumes a counterfactual scenario, and thus cannot be fully satisfied. Yet it can be approximately satisfied. A metrological *realization* is a system that approximately satisfies the definition of the measurand. Realizations are used to operationalize the definition of the measurand, so as to make its value (or values) empirically accessible. For example, there are currently over a dozen primary frequency standards operational around the world. These are atomic clocks that serve as the most accurate measurement standards for time and frequency metrology. Each of these clocks measures the radiation frequency associated with cesium-133 atoms under conditions that closely approximate the ideal conditions specified by the definition of the second. However, no approximation is perfect, and different realizations deviate from the ideal in different respects and degrees. Consequently, metrologists do not consider any of the primary frequency standards to be completely accurate. Doing so would lead to inconsistencies, as the clocks 'tick' at slightly different rates due to differences in the conditions affecting the cesium atoms in each laboratory. Instead, metrologists develop detailed theoretical and statistical models of each clock, and test these models by experimenting on the clocks and measuring their surrounding environment [18,23]. These models are then used to estimate the deviation of each clock from the ideally defined frequency [41].

When a less accurate clock is calibrated against a primary frequency standard, the accuracy of the clock is not evaluated simply by its ability to reproduce the frequency of the primary standard. Doing so would make the less accurate clock inherit the frequency biases of the standard. Instead, the biases and uncertainties associated with the primary standard are included in the accuracy evaluation of the less accurate clock [33]. This procedure ensures that accuracy is evaluated relative to the theoretical *definition* of the second, rather than against any of its idiosyncratic, concrete realizations. By following this procedure, clocks that were calibrated against different primary realizations of the second provide consistent estimates of time and frequency, even though the raw frequencies ('tick' rates) of primary realizations disagree with each other.

Metrological accuracy is the closeness between the measured quantity value and the value of the measurand as defined. The defined value of the measurand is considered to be unknowable, and only capable of approximation with some uncertainty. Much of the conceptual and mathematical apparatus of metrology is dedicated to estimating bias, understood as a systematic difference between measured and defined quantity values. As the defined value of the measurand is unknowable, estimations of bias are necessarily inexact and involve some uncertainty. To evaluate this uncertainty, metrologists assess the extent of deviation between the actual operating conditions of their instruments and the ideal operating conditions specified by the definition of the measurand.

Metrologists employ a variety of strategies to acquire this counterfactual information. Some of these strategies involve physically controlling elements of the apparatus and environment so that they more closely approximate the ideal, e.g., controlling the temperature of the environment. But these physical control strategies ultimately reach a practical limit. To go beyond this limit, metrologists use a combination of theoretical predictions and secondary experiments to assess how the apparatus *would have* behaved if its operating conditions were closer to the ideal. For example, they vary the density of cesium atoms in the atomic clock, and use theory and statistics to predict what the frequency of the clock would have been at zero density [17:328–9]. The uncertainty associated with this prediction becomes a component of the measurement uncertainty of the clock.

The upshot is that a clock's accuracy is a property of a *predictive inference* [42]. Accuracy ultimately depends on the ability of scientists to use the clock's indications ('ticks') to predict the value of a latent, counterfactually defined frequency. The accuracy of this prediction, and therefore of the clock itself, depends on extensive and domain-specific background knowledge, and cannot be reduced to an association or matching between observations.

## 5   Target specification bias and its implications for fairness

The discussion of time and frequency metrology provided above leaves out much detail, but even this cursory survey suggests similarities between the inferential structures of measurement and supervised ML-based prediction. Both are types of method for evaluating variables based on concrete input (whether a new example, or an object to be measured). Both involve a modeling (training or calibration) phase, in which reliable data (training dataset, or values associated with standards) are used to generate stable associations between the inputs and outputs of the instrument [31]. In both cases, the associations revealed during the modeling phase are generalized to new objects or events in the application (deployment or measurement) phase. In both cases, the model is optimized to increase the accuracy of predictions of values of the target variable (measurand). Finally, both types of method presume to provide evidence for decision making, and are often presented to decision makers as trustworthy within reasonable limits.

These similarities are perhaps not surprising, given that both measurement and predictive ML rely on inductive reasoning. I do not wish to overemphasize the similarities: there are many dissimilarities between measurement and predictive ML as well. These include different modes of implementation (computational versus material), the fact that the input of ML is a representation rather than a concrete object or event, and the fact that an ML model is a model of the *data* rather than a model of the measurement process. There are also many methodological and institutional dissimilarities. Yet the similarities in inferential structure are sufficient to support a reasonable hope that some helpful lessons may be drawn from metrology, which is a significantly older and more methodologically mature field than ML research, for tackling current challenges facing ML research. In what follows I will focus on four such lessons.

### 5.1   Lesson One: labels are not intended to reflect target variables, but to operationalize them

From an abstract, mathematical perspective, ML may be viewed as no more than a 'regression machine' that fits a function to data under specified constraints. However, the practical problems that ML tools are commonly deployed to solve, such as optimizing resource allocation or predicting the occurrence of a disease, are not identical to the regression problems that ML tools are designed to solve. Rather, the regression problem meant to be solved by a given ML tool is an *operationalization* of the real-world problem. Even if the model is a good solution to the regression problem specified by algorithm designers, it does not yet follow that the model is a good solution to the real-world problem that the model will be deployed to solve. This is a familiar situation in measurement: a measuring instrument almost never measures precisely the same variable that users are interested in measuring.

The difference between the target variable and the variable being measured is often subtle. Unless the target variable is carefully defined, the discrepancy may go unnoticed. In some cases, the discrepancy may be practically negligible, while in others, an unnoticed discrepancy can entail significant harm. To avoid such harms, the real-world problem and its operationalization need to be clearly distinguished from each other, and their relationship carefully studied.

Metrologists are used to asking themselves whether the target variable as defined is identical to what the instrument is designed to measure, and whether the instrument *in fact* measures what it is designed to measure. Most commonly, the answer to both questions is 'no'. Similarly, designers of predictive ML tools typically benefit from asking: (i) What variable do labels in the actual data reflect? (ii) What variable are labels *intended* to reflect? and (iii) How does the variable that labels are intended to reflect differ from the target variable as defined by stakeholders?

Questions (i) and (ii) are increasingly at the center of attention in ML research, as evidenced by recent work on measurement error, label bias, and biased proxy variables [24,30]. The third question is more seldomly raised, perhaps due to the belief that the variable the labels are intended to reflect is identical to the variable of interest to stakeholders. But in well-designed predictive tools, these two variables should usually be distinct. As already mentioned, the variable of interest to stakeholders is typically not practically realizable even in principle. Much like the frequency of an ideal, unperturbed cesium atom, decision makers are typically interested in counterfactually defined variables, such as the *ceteris paribus* prognosis of patients under equal treatment, or the *ceteris paribus* health risks to a patient if diagnostic testing were not selective. Even in the best practically possible data acquisition scenario, labels that reflect the variables of interest to stakeholders are not attainable, because the conditions the define such variables are never fulfilled. Intending labels to directly reflect target variables ignores the inferential gap between the two, with potentially harmful and unjust consequences to patients.

Instead, fairness and safety are better served by viewing intended labels as operationalizations that necessary satisfy the definition of the target variable only approximately, and to account for this approximation when reporting accuracy to stakeholders. Metrologists are already taking this responsible approach to accuracy evaluation and reporting: they reconcile multiple, idiosyncratic measuring instruments by accounting for their deviations from a common, counterfactual ideal. The mutual reconciliation of measurements results relative to a counterfactual ideal is essential to their reproducibility, and a precondition for many practical applications, including precision timekeeping, modern manufacturing, and reliable communications.

## 5.2    Lesson Two: target specification bias is distinct from data acquisition error and label bias

Operationalizing a target variable is a complex activity. It involves an iterative investigation of the degree of alignment between the goals of variable estimation and the design of concrete estimation procedures. The result is often a chain, or a hierarchy, of variables starting with a highly idealized variable definition, proceeding through successive approximations, and terminating with the results of one or more concrete methods. For example, when measuring mass in kilograms, the idealized variable is a ratio between the mass of an object and a defined physical constant. The physical constant that currently defines the kilogram is a function of the Planck constant, the speed of light in vacuum, and the ideal cesium transition frequency [3]. If everyday kitchen scales were required to directly estimate this ratio, they would have been tremendously complex and expensive. Instead, kitchen scales are calibrated to a variable that is located at the bottom of a long chain of operationalizations, namely to the mass of a standard metal weight that approximates the defined constant. Operationalization is successful, not when these variables are identical, but when the relationships among them are stable and sufficiently well-known.

Different links in the chain of operationalizations introduce different kinds of error. This holds true for measurement as it does for the operationalization of variables in ML-based decision support tools. A typical case of supervised ML is illustrated in Figure 2. Errors that originate at the lowest levels of the chain – the level of data acquisition – are commonly known as 'measurement errors' or 'measurement biases' [30:5]. These include, for example, incorrect or missing data, noise in the data, and duplicate records. Such errors introduce bias into the relationship between labels in the training data and the events or objects in the world that the labels are meant to represent. For example, an error in the recording of a diagnostic test could cause a positive test result to be recorded as negative, or vice versa. Next comes 'label bias', namely errors that arise due to differences between the intended and actual labels. Labels are usually intended to reflect some actual state of affairs in the world, such as the existence of a pathology. Actual labels are indirect approximations of intended ones. A common source of label bias is error in the underlying evidence-collection process. For example, the diagnostic test itself may be inaccurate, e.g., a pathology exists but is not detected. Even if the diagnostic test is completely accurate, label bias could still arise if, for example, the sample on which data is collected is not representative of the intended population, e.g., the sample contains significantly more men than women or an unbalanced age distribution relative to the population of interest.

Target specification bias is distinct from data acquisition error and label bias. This kind of bias arises due to differences between the intended labels and the target variable. As mentioned, in decision-making scenarios target variables are typically counterfactually defined, and concern a state of affairs where confounders are absent. It is often practically impossible to remove confounding factors by directly intervening in the world. For example, it is impossible to remove systematic health inequalities from society before collecting the data. Even when direct elimination of confounders is practically possible, it may not be ethically or legally permissible. For example, doing so in the pneumonia case would require withholding special treatment from asthmatics who present pneumonia symptoms. Consequently, target variables must usually be defined in a counterfactual world.

The counterfactual nature of target variables distinguishes them from intended labels. Labels are data about the actual, rather than a counterfactual, world. For example, they are intended to reflect the actual health outcomes of pneumonia patients in a manner that is free from inaccuracies, i.e., free from data acquisition error and label bias. At the same time, labels are *not* intended to reflect a counterfactual world free of upstream decision making (such as differential treatment of at-risk patients), diagnostic suspicion bias (such as unjustified differences in diagnostic procedure based on patients' race, gender, or age) or systematic injustices (such as unequal access to healthcare). No method can directly collect data about such counterfactual worlds, because they are not empirically accessible. Properties of such counterfactual worlds – including the target variable – must be inferred from data about the actual world based on some assumptions. The nature of these assumptions will be discussed under Lesson Four.

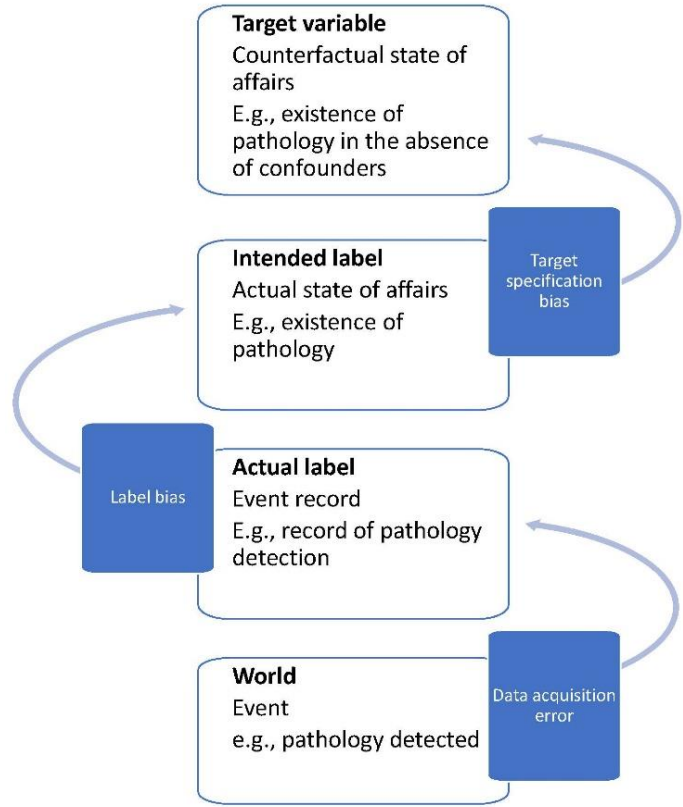

**Figure 2: A typical chain of operationalizations of a target variable in supervised ML for healthcare decision making. Target specification bias arises due to differences between intended labels and the target variable, and is distinct from label bias and data acquisition error.**

It should be noted that target specification bias is a broader category than proxy bias, at least under a common interpretation of the term 'proxy'. Proxy bias arises when the labels and the target variable are defined as different *kinds* of properties, such as when the cost of healthcare serves as a stand-in for the severity of illness. The difference between the target variable and its operationalization is especially stark in such cases. Nonetheless, target specification bias persists even when the two variables are of the same kind, such as when actual severity of illness is used to estimate counterfactual severity of illness in the absence of confounders. Consequently, unlike proxy bias, target specification bias cannot be completely solved simply by picking better labels [34:5]. Of course, one could decide to use the term 'proxy' very broadly and apply it to any operationalization. Broadly speaking, all measurement and supervised ML involve proxy variables, because the variable of interest is never completely identical to its operationalization. However, such linguistic usage would obscure the specific challenges presented by what ML researchers typically call 'proxy variables' and the special solutions available to address these challenges.

## 5.3    Lesson Three: target specification bias affects both accuracy and fairness

If not corrected, target specification bias diminishes model accuracy. To see how, one must first consider the concept of accuracy itself. On a narrow, label-matching concept of accuracy, the optimal targets of prediction in supervised ML are actual states of affairs, such as the actual rates of patient recovery and mortality. Accordingly, when a model is optimized in a manner that sacrifices its degree of fit with labels – for example, to accommodate fairness constraints – its predictions are viewed as less accurate than they would be in the absence of such constraints. This conclusion is consistent with the claim, commonly made by ML researchers, that accuracy and fairness trade-off against each other [11,35].

On a closer look, there are at least two distinct concepts of benchmark accuracy to consider when evaluating the performance of ML-enabled decision support tools. The first is the accuracy of labels relative to the actual states of affairs that labels are intended to describe, e.g., whether a patient that is recorded as having survived pneumonia in fact survived. Here the benchmark is the intended label. This is the benchmark commonly used under the label-matching concept of accuracy. The second concept of benchmark accuracy is the accuracy of labels relative to the variable decision makers are interested in predicting, e.g., whether a patient that is recorded as having survived pneumonia would have survived if all patients had received the same treatment. Here the benchmark is the target variable. Under this second, broader concept of benchmark accuracy, even the most reliably acquired labels are still imperfect operationalizations of the target variable. By analogy, even a physical clock that works precisely as intended would still be affected by confounders that make it an imperfect operationalization of the theoretical definition of the standard SI second.

It is this broader conception of accuracy that usually interests decision makers. Under this conception, accuracy is evaluated relative to a benchmark that is meaningful to decision makers, i.e., that represents the kind of evidence decision makers are seeking, rather than a technical aspect of the algorithm's validation and testing. The upshot is that a good fit to the labels – even once the labels are corrected for data acquisition errors and label bias – may not be sufficient to guarantee accuracy from the perspective of decision makers. Indeed, in some cases a good fit between model predictions and corrected labels may be a sign of *inaccuracy*. This is especially the case if there are reasons to think that the actual world in which the labels were collected differs substantially from the counterfactual world about which decision makers seek evidence. Evaluating model accuracy relative to a counterfactually specified target variable takes target specification bias into account, resulting in more complete and user-relevant accuracy estimates than those based strictly on label-matching.

Target specification bias also diminishes the fairness of decisions that are based on model predictions. From the perspective of a decision maker who is interested in making fair decisions, upstream medical decisions that affect the distribution of health outcomes across groups are confounders. This is the case regardless of whether those decisions are unjust (e.g., due to health disparities) or due to justified differential treatment (e.g., preferential treatment to asthmatics). The target variable is specified in the absence of such confounders, on a counterfactual world that is free from differential intervention. This counterfactual approach to defining target variables does not free decision makers from addressing difficult theoretical questions about what exactly they mean by 'fairness'. On the contrary, the emphasis on target specification as a distinct, theoretical task that involves societal and ethical considerations highlights the potential conflicts among different conceptions of fairness.

Fairness criteria that are incorporated into the definition of the target variable become part of the accuracy benchmark for the relevant decision support tool. Under a broad, user-oriented conception of accuracy, implementing such fairness criteria does not trade off against accuracy, but rather aligns with the aim of improving model accuracy. For example, correcting the predictions of the pneumonia hospitalization decision support tool so that asthmatics are prioritized (rather than de-prioritized) increases both accuracy and fairness. Accuracy is increased by providing decision makers with evaluations of the target variable they are interested in – in this case, how asthmatics would fare in the absence of differential treatment – and fairness is increased by better aligning resource allocation with medical need. Target specification bias therefore defies the typical trade-off between fairness and accuracy.

## 5.4    Lesson Four: mitigating target specification bias requires domain-specific knowledge

In the analogy with measurement, target specification bias is a type of systematic measurement error. Unlike random error, systematic error is a "component of measurement error that in replicate measurements remains constant or varies in a predictable manner" [22:2.17]. Systematic errors often stem from background processes and assumptions that remain stable when the measurement is repeated, such as a background gravitational field or a biased estimation of a physical constant. Such errors cannot be detected by applying statistical tests to repeated measurements, but must be inferred from theoretical models of the measurement process, by performing additional measurements, or by using established measurement standards.

Similarly, target specification bias stems from processes and assumptions that remain stable when the same part of the world is resampled. Gathering additional data from the same hospital would not reveal that the model underestimates the risk of pneumonia to asthmatics, because the process of differential treatment that gives rise to the bias remains constant. Nor would the bias be revealed by using different labels, employing different measures of fit between predictions and labels, or employing a generic fairness criterion that equalizes some performance parameter across patient groups. Recall that the performance parameter that decision makers are interested in equalizing is defined counterfactually: it is the allocation of resources by health risk when all other things are equal. The relevant sense of 'all other things' is domain-specific, and depends on the context of the decision at hand. Mitigating target specification bias requires decision makers and algorithm designers to explicitly specify their assumptions concerning what needs to 'remain equal' in the counterfactual scenario. Then, they must formulate and empirically test hypotheses concerning the differences between this counterfactual scenario and the actual one. In doing so, they would be following the example of metrologists, who theorize about the deviations of their clocks from the idealized definition of the SI second.

This is not to imply that collecting and analyzing data cannot help to mitigate target specification bias. One way of formulating hypotheses about the actual processes that give rise to data is to increase the transparency of the model, so that the correlations it discovers become more easily surveyable. In the case of Caruana's model, this was achieved by training a rule-based learning algorithm on the same data. Another helpful family of techniques employ methods of causal inference that reveal counterfactual probabilities in the data. For example, the use of Bayesian networks or structural equation models can reveal causal dependencies that are relevant to healthcare decision making [36]. Recent work in explainable AI (XAI) has featured breakthroughs in extracting counterfactual information from ML models and presenting it to users, with potential applications for clinical decision support tools [7,38].

Such methods should be used in combination with clinical judgment to interpret the resulting counterfactuals and determine which of them is relevant for the decision at hand. Importantly, decision makers need to exercise judgment when deciding which counterfactual conditions need to be equalized across which patient groups. For example, it makes little sense to allocate medical resources to children suffering from asthma based on the diagnosis they would have received if they were adults. The medical resources, diagnostic criteria, and treatment options are far too different between these two groups to make such counterfactual information relevant for decision making. Domain-specific and contextual knowledge remains crucial for specifying which counterfactual information is relevant for a given type of medical decision. This point is further strengthened by the fact that various stakeholders, including patients, physicians, healthcare administrators, insurers, and health policymakers may have conflicting specifications for the same target variable. In such cases, addressing target specification bias requires an inclusive consultation regarding the precise aims of prediction.

# 6  Conclusion: towards a metrological evaluation of accuracy for machine learning

With the proliferation of ML-based tools into areas of high-stakes decision making, such as healthcare, criminal justice, finance, and defense, the methods used to evaluate and report the accuracy of ML models need to conform to stringent standards. The discussion above suggests that the label-matching conception of predictive accuracy is inadequate and potentially harmful for supporting high-stakes decisions. It is misleading to report the rate of label-matching to stakeholders (whether in terms of sensitivity, specificity, AUC, or some other metric) and present it as the ultimate evaluation of the model's accuracy, even if the labels themselves are highly reliable. Instead, the accuracy of decision support tools should be reported relative to a counterfactually defined target variable, with an uncertainty margin that reflects the unknown degree of error around reported values. Methods for reporting this sort of counterfactual information are still in their infancy in ML [43], but are highly developed in metrology, which could serve as a role model for future developments.

Label-matching metrics of accuracy may still be useful for internal model validation, including testing for under- and over-fitting of the model. Such metrics can reflect the internal validity of the model, that is, an evaluation of the fit between the associations learned by the model and the associations found in the part of the world from which data was collected. Such metrics express how well model predictions generalize from the training dataset to the test dataset, and are therefore tied to the idiosyncratic conditions under which these datasets were obtained. This sort of generalizability may be sufficient for some low-stakes decisions, such as retail consumer purchasing recommendations, but not for medical decision making, which is subject to a higher standard of harm prevention and requires a systematic exclusion of confounders. Rigorous, metrological accuracy evaluation of ML decision support tools will have the benefits of reducing target specification bias, providing clearer and more actionable information to users, increasing fairness, and improving reproducibility and public trust.

**Acknowledgments**

I am grateful to Yasmin Haddad, Ljubomir Raicevic, Branden Fitelson, and Momin Malik for our discussions, to three anonymous reviewers for their comments, and to audiences at Northeastern University, University of Oregon, University of Guelph, Leibniz University Hannover, University of Memphis, and Université du Québec à Montréal for their feedback. Funding for this research was provided via the Canada Research Chairs Program (grant CRC-2019-00119).